\documentclass[nohyperref]{article}

\usepackage{microtype}
\usepackage{graphicx}
\usepackage{subfigure}
\usepackage{booktabs} 

\usepackage{hyperref}

\usepackage[accepted]{icml2022}

\usepackage{amsmath}
\usepackage{amssymb}
\usepackage{mathtools}
\usepackage{amsthm}

\usepackage{wrapfig}
\usepackage{mathtools, cuted}

\usepackage[capitalize,noabbrev]{cleveref}

\DeclareMathOperator*{\argmax}{arg\,max}
\DeclareMathOperator*{\argmin}{arg\,min}
\theoremstyle{plain}

\theoremstyle{definition}

\theoremstyle{remark}

\usepackage[textsize=tiny]{todonotes}

\icmltitlerunning{Bit-level Inference Scaling Laws}

\begin{document}

\twocolumn[
\icmltitle{The case for 4-bit precision: \\k-bit Inference Scaling Laws}



\icmlsetsymbol{equal}{*}

\begin{icmlauthorlist}
\icmlauthor{Tim Dettmers}{yyy}
\icmlauthor{Luke Zettlemoyer}{yyy}
\end{icmlauthorlist}

\icmlaffiliation{yyy}{University of Washington}

\icmlcorrespondingauthor{}{dettmers@cs.washington.edu}

\icmlkeywords{Machine Learning, ICML}

\vskip 0.3in
]



\printAffiliationsAndNotice{} 

\begin{abstract}
Quantization methods reduce the number of bits required to represent each parameter in a model, trading accuracy for smaller memory footprints and inference latencies. However, the final model size depends on both the number of parameters of the original model and the rate of compression. For example, a 30B 8-bit model and a 60B 4-bit model have the same number of bits but may have very different zero-shot accuracies. In this work, we study this trade-off by developing inference scaling laws of zero-shot performance in Large Language Models (LLMs) to determine the bit-precision and model size that maximizes zero-shot performance. We run more than 35,000 experiments with 16-bit inputs and k-bit parameters to examine which zero-shot quantization methods improve scaling for 3 to 8-bit precision at scales of 19M to 176B parameters across the LLM families BLOOM, OPT, NeoX/Pythia, and GPT-2. We find that it is challenging to improve the bit-level scaling trade-off, with the only improvements being the use of a small block size -- splitting the parameters into small independently quantized blocks -- and the quantization data type being used (e.g., Int vs Float). Overall, our findings show that {4-bit} precision is almost universally optimal for total model bits and zero-shot accuracy.
\end{abstract}

\section{Introduction}

Large Language Models (LLMs) are widely adopted for zero/few-shot inference~\cite{zhang2022opt,black2022gpt,zeng2022glm,scao2022bloom}, but they can be challenging to use both due to their large memory footprints -- up to 352 GB of GPU memory for 175B models -- and high latency. However, both the memory and latency are primarily determined by the total number of bits in the parameters. Therefore, if we reduce the model bits through quantization, we can expect the latency of the model to reduce proportionally, potentially at the expense of end task accuracy~\citep{frantar2022gptq,park2022nuqmm,yao2022zeroquant}.

\begin{figure}[t]
     \centering
         \includegraphics[scale=0.55]{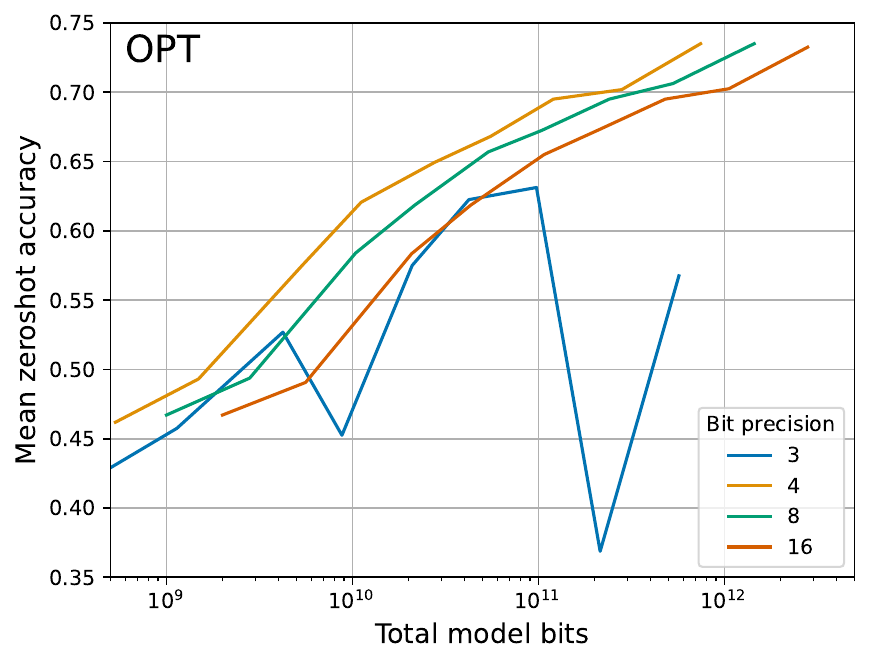}\vspace{-1.0em}
        \caption{Bit-level scaling laws for mean zero-shot performance across four datasets for 125M to 176B parameter OPT models. Zero-shot performance increases steadily for fixed model bits as we reduce the quantization precision from 16 to 4 bits. At 3-bits, this relationship reverses, making 4-bit precision optimal.\vspace{-1.5em}}
        \label{fig:opt}
\end{figure}

Since we can quantize the parameters of a trained model to an arbitrary bit-precision, this raises the question of how many bits should be used to optimally trade off accuracy and total model bits, given our current base methods for model quantization. For example, {\bf if we have a 60B LLM in 4-bit precision and a 30B LLM in 8-bit precision, which will achieve higher accuracy?} To study such trade-offs, it is helpful to take the perspective of scaling laws \citep{kaplan2020scaling,henighan2020scaling}, which evaluate the underlying trends of variables to make generalizations beyond individual data points.

In this paper, we study bit-level inference scaling laws for zero-shot quantization to determine the precision that maximizes zero-shot accuracy given a certain number of total bits in the model. {\bf Our main finding is that 4-bit parameters yield optimal performance for a fixed number of model bits across all model scales and model families tested}. 

We study five different model families, OPT, Pythia/NeoX, GPT-2, BLOOM, and BLOOMZ \citep{zhang2022opt,black2022gpt,radford2019language,scao2022bloom}, with 19M to 176B parameters, and for 3 to 16-bit precision. In addition, we run more than 35,000 zero-shot experiments to vary many of the recently studied advances in quantization precision, including the underlying data types and quantization block size.
We find that reducing the precision steadily from 16 to 4 bits increases the zero-shot performance for a fixed number of model bits, while at 3-bit, the zero-shot performance decreases. This relationship holds across all models studied and did not change when scaling from 19M to 176B parameters, making 4-bit quantization universally optimal across all cases tested.

Beyond this, we analyze which quantization methods improve and which degrade bit-level scaling. We find that none of the quantization methods we test improve scaling behavior for 6 to 8-bit precision. For 4-bit precision, data types and a small quantization block size are the best ways to enhance bit-level scaling trends. We find that quantile quantization and floating point data types are the most effective, while integer and dynamic exponent data types generally yield worse scaling trends. For most 4-bit models, a block size between 64 to 128 is optimal in our study.

From our results, we can make straightforward {\bf recommendations for using zero-shot quantized models for inference}: always use 4-bit models with a small block size and with a float data type. If trade-offs in total model bits or predictive performance are desired, keep the precision in 4-bit but vary the number of parameters of the model instead.

While earlier work has shown that it is possible to significantly improve the predictive performance of quantized models by using outlier-dependent quantization \citep{dettmers2022llm,xiao2022smoothquant}, we show that this is not effective for bit-level scaling. In particular, we analyze instabilities in 3-bit OPT and Pythia models and show that while these models can be stabilized through an outlier-dependent quantization which we term {\it proxy quantization}, this does not improve bit-level scaling compared to 4-bit precision. On the other hand, we highlight that one-shot quantization methods, that is, methods that optimize the quantization through a single mini-batch of data, potentially could be scaled to be bit-efficient below the 4-bit level.
Overall, these findings suggest that the most promising directions to improve zero-shot bit-level scaling laws are to develop new data types and techniques that quantize outliers with high precision without requiring a significant amount of additional bits. We also highlight the potential for one-shot quantization methods as a way towards low-bit transformers if combined with our insights.

\section{Background}

It might be unintuitive that reducing the number of bits of a model is directly related to inference latency for LLMs. The following section gives the background to understand this relationship. Afterward, we provide a background on quantization data types and methods.

\subsection{Relationship between inference latency and total model bits}
\label{lbl:inference-latency-bits}

While the main goal of our work is to find the best trade-offs between model bits and zero-shot accuracy for LLMs, the total model bits are also strongly related to inference latency. The overall computation latency -- the time it takes from start to finish of a computation -- is mainly determined by two factors: (1) how long does it take to load the data from main memory into caches and registers, (2) how long does it take to perform the computation. For example, for modern hardware like GPUs, it usually takes more than 100 times longer to load a number than to do an arithmetic operation with that number \citep{jia2019dissecting, dongarra2022}. Therefore, reducing the time spent loading data from main memory is often the best way to accelerate overall computation latency. Such reductions can be achieved mainly through caching and lower precision numbers.

Caching can reduce the overall latency of matrix multiplication by a factor of 10x or more \citep{jia2019dissecting}, given that neither matrix entirely fits into the L1 cache of the device. In matrix multiplication, ${\mathbf{b}\mathbf{W}=\mathbf{h}}$, with the batch $\mathbf{b}$ and parameter matrix $\mathbf{W}$, we load each row of $\mathbf{b}$ and each column of $\mathbf{W}$ and multiply them -- thus multiple loads from global memory occur for each row/column which can be cached. If ${\mathbf{b}}$ entirely fits into the L1 cache, no reuse is possible because neither $\mathbf{W}$ nor ${\mathbf{b}}$ is loaded more than once from global memory. This occurs if the inference batch size is below 60 or 200 for an RTX 3090 or RTX 4090 GPU. As such, caching is ineffective below these batch sizes, and the inference latency is solely determined by the memory loaded from $\mathbf{W}$.

Thus in the case where the mini-batch fits into the L1 cache, inference latency can be reduced by using a smaller model with smaller $\mathbf{W}$ or by compressing an existing model to use a lower bit-precision per parameter in $\mathbf{W}$. For example, beyond their algorithmic innovation of improved rounding for quantization, \citet{frantar2022gptq} also developed inference CUDA kernels for 16-bit inputs and 3-bit integer weights, which yields inference latency improvements of up to 4.46x compared to 16-bit inputs and weights for OPT-175B -- close to the 5.33x reduction in model bits. As such, reduction in the total model bits is strongly correlated with inference latency for small inference batch sizes.

\subsection{Data types}

Here we provide a brief overview of the data types that we study. Please see Appendix~\ref{appendix:datatypes} for full specification of these data types.

We use four different data types. For \textbf{Integer} and \textbf{Float} data types, use IEEE standards. Our Float data type has an exponent bias of $2^{E-1},$ where $E$ is the number of exponent bits. We also use \textbf{quantile quantization}, a lossy maximum entropy quantization data type \citep{dettmers2022optimizers}, where each quantization bin holds an equal number of values. This ensures that each bit pattern occurs equally often. Finally, we use \textbf{dynamic exponent quantization} \citep{dettmers20168bit}, which uses an indicator bit to separate an exponent bit region and a linear quantization region. The exponent can vary from value to value by shifting the indicator bit. This data type has low quantization error for tensors which have numbers that vary by many orders of magnitude. 

\subsection{Blocking / grouping }

Quantization precision is, in part, determined by whether all quantization bins are used equally. For example, a 4-bit data type has 16 bins, but if, on average, only 8 bins are used, it is equivalent to a 3-bit data type. As such, methods that help to increase the average use of all quantization bins in the data type can increase quantization precision. In this subsection, we introduce blocking. Blocking/grouping methods chunk the tensor into smaller pieces and quantize each block independently. This helps to confine outliers to particular blocks, which increases the average number of bins used across other blocks and thus increases the average quantization precision.

\paragraph{Blocking/Grouping.} Blocking and grouping are similar concepts. In grouping, we sub-divide a tensor along a certain dimension into $n$ parts and assign each sub-tensor, called group, its own normalization constant $c$, which is usually the absolute maximum of the group. In blocking, we view the tensor as a one-dimensional sequence of values and divide this sequence into parts of size $n$ called blocks.

In our work, we use blocking because, unlike grouping, it provides a measure of additional bits per parameter independent of the hidden dimension. For example, using 16-bit normalization constants and a block size of 64 means, we have an extra 16 bits every 64 parameters or 16/64=0.25 bit-per-parameter additional cost for using block-wise quantization -- this is true for every model regardless of hidden dimension. For grouping, the exact cost would depend on the size of the hidden dimension of each model. 

For block-wise quantization, we use the notation of \citet{dettmers2022optimizers}, which defines the block-wise quantization with $k$ bits, block-size $B$, input tensor $\mathbf{T}$ with $n$ elements, $n/B$ blocks as follows. If $\mathbf{{Q}}_k^{\text{map}}(\cdot)$ maps the integer representation of a data type to the representative floating point value, for example, the bit representation of a 32-bit float to its real value, and if we define the index of each block in $0..n/B$ by index $b$, and we compute the normalization constant as $m_b = \max(|\mathbf{T}_b|)$, then block-wise quantization can be defined by finding the minimum distance to the value of the quantization map as follows:

\begin{equation}
    \mathbf{T}_{bi}^{Q_k^{\text{map}}} = \argmin\limits_{j=0}^{n=2^k} |\mathbf{{Q}}_k^{\text{map}}(j) - \frac{\mathbf{T}_{bi}}{m_b}|{\bigg\rvert}_{0<i<B},
\end{equation}

\section{Outlier-dependent quantization through proxy quantization}

Outlier features that emerge in large language models \citep{gao2019representation, timkey2021all,bondarenko2021understanding,wei2022outlier,luo-etal-2021-positional,kovaleva2021bert,puccetti2022outliers} can cause large quantization errors and severe performance degradation \citep{dettmers2022llm,zeng2022glm,xiao2022smoothquant}. While it has been shown that it is sufficient to use 16-bit inputs and 8-bit weights to avoid this disruption \citep{zeng2022glm}, it is unclear if outlier features can cause degradation if we use 16-bit inputs and weights below 8-bit precision.

To this end, we develop outlier-dependent quantization through proxy quantization, where we quantize weights to a higher precision for the corresponding outlier feature dimensions to test how much precision is needed for the weights.
A significant challenge is that each model has a different number of outlier features, and outliers partially depend on inputs that are different for each zero-shot task. As such, we seek a model-independent model that has a constant memory footprint across all models and tasks.

In initial experiments, we noted that the criterion developed by \citet{dettmers2022llm}, which thresholds the hidden states to detect outlier features, is unreliable as it depends on the standard deviation of the hidden state. This causes problems because, for models such as OPT, the standard deviation of the hidden states increases in later layers, which causes too many outliers to be detected. This also has been noted by \citet{zeng2022glm}.
By inspecting this anomaly in OPT models, we find that a better measure of detecting outlier dimensions is the standard deviation of the weights of each hidden unit of the previous layer. The standard deviation of hidden unit weights that produce outliers are up to 20x larger than the standard deviation of other dimensions. 

With this insight, we develop what we call \textbf{proxy quantization}. Proxy quantization is input-independent and, therefore task-independent, as it uses the standard deviation of each layer's hidden unit weights as a proxy for which dimensions have outlier features. For example, given a transformer with $n$ linear layers (FFN and attention projection layers) with weight matrices $\mathbf{W}_i\in \mathbb{R}^{h\times o}$ where $h$ is the input dimension and $o$ the output dimension (thus $o$ hidden units), we define the set of indices $J$ to be quantized in higher precision by:
\begin{equation}
    J_{i+1} = \argmax\limits_{j=0}^k\text{std}(\mathbf{W}_i){\big\rvert}_{i..n}
\end{equation}

where std$(\cdot)$ is the standard deviation of the output dimension $o$. We then quantize the input dimensions of the weight of the next layer in 16-bit precision if it is in set $J$ and k-bit otherwise.

\section{Experimental setup}

In our experiments, we use 16-bit inputs and k-bit quantized parameters for $3 \geq k \geq 8$. Attention matrices are not quantized since they do not contain parameters. We also use a 16-bit baseline that does not use any quantization (16-bit floats).

To measure inference performance for k-bit quantization methods, we use perplexity on the CommonCrawl subset of The Pile \citep{gao2020pile} and mean zero-shot performance on the EleutherAI LM Evaluation harness \citep{eval-harness}. In particular, for the zero-shot setting, we use the EleutherAI LM eval harness \citep{eval-harness} in the GPT-2 setting on the tasks LAMBADA \citep{paperno-etal-2016-lambada}, Winogrande \citep{DBLP:journals/cacm/winogrande2021}, HellaSwag \citep{DBLP:conf/acl/hellaswag2019}, and PiQA \citep{DBLP:conf/aaai/piqa2020}.

The choice of these particular zero-shot tasks was mainly motivated by previous work \citep{dettmers2022llm,yao2022zeroquant,xiao2022smoothquant}. However, in our evaluation, we find that perplexity is a superior metric since its continuous value per sample leads to less noisy evaluations. This has also been noted by \citet{frantar2022gptq}. For example, when using perplexity to evaluate data types, quantile quantization is the best data type. Still, when we use zero-shot accuracy as an evaluation metric, the float data type is sometimes better because zero-shot accuracy is noisier. Furthermore, we find that across more than 35,000 zero-shot experiments, the {\bf Pearson correlation coefficient between The Pile Common Crawl perplexity and zero-shot performance is -0.94}. 

This highlights that perplexity is sufficient and preferable for evaluation purposes. A serious drawback is that perplexity is challenging to interpret. As such, we use zero-shot accuracies in the main paper for clarity but encourage the reader to use perplexity measures found in the appendix for replication and comparison purposes. Since perplexity evaluations are so reliable, it is possible to replicate our work by evaluating a small number of samples using the perplexity metric, which makes the construction of scaling laws less computationally intensive.

\textbf{Scaling laws.} We try to fit power laws to our data, but we find that bivariate power functions with respect to the number of parameters and the bit-precision provide a poor fit. However, when we fit linear interpolations to represent scaling curves, we find that different bit-precisions are almost parallel, indicating that the scaling trends of different precisions can be represented faithfully by a base function and an offset for each bit-precision. As such, we choose to use linear interpolations to represent scaling trends.

\begin{figure*}[t]
     \centering
         \includegraphics[scale=0.55]{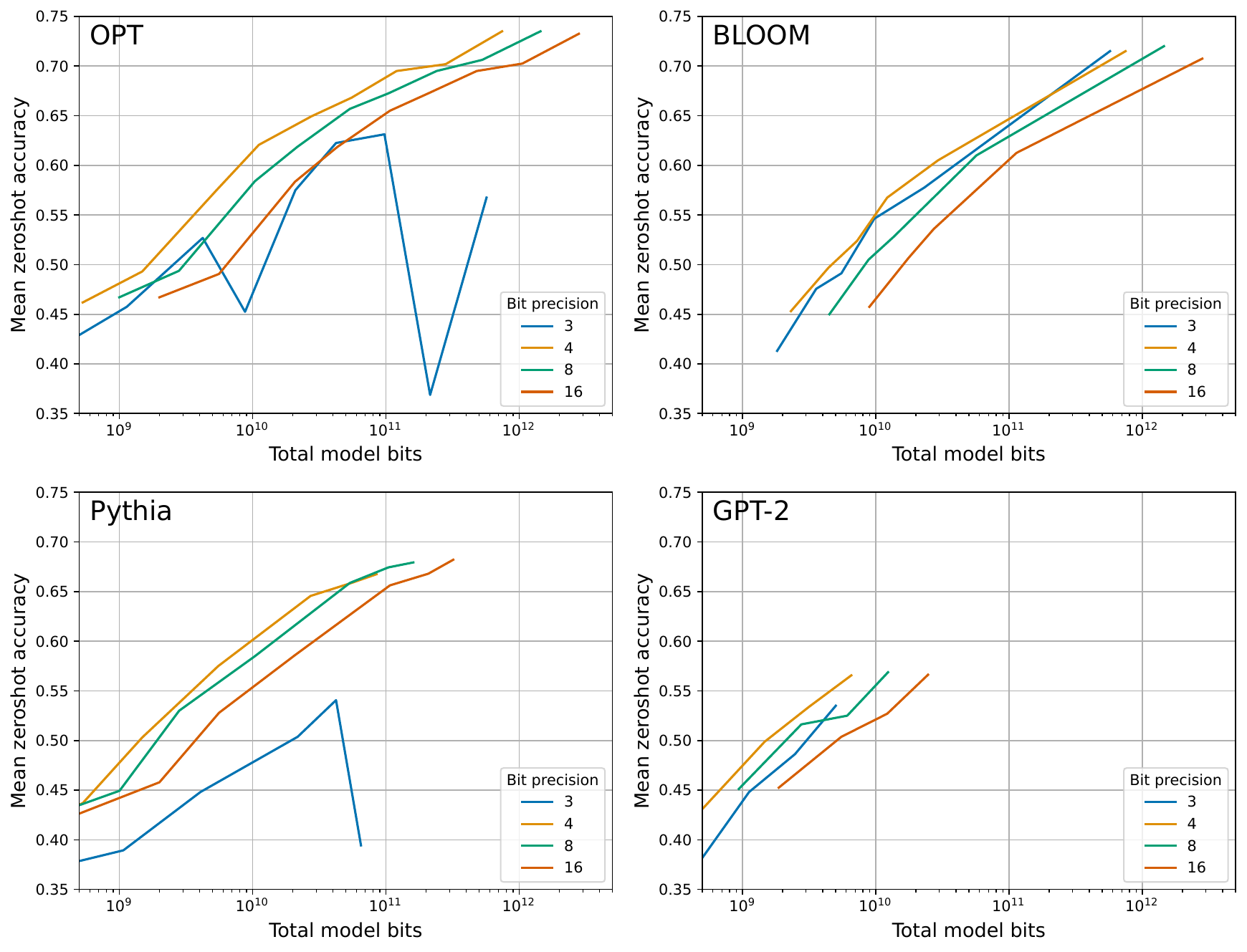}
        \caption{Bit-level scaling laws for mean zero-shot performance across Lambada, PiQA, Winogrande, and Hellaswag. 4-bit precision is optimal for almost all models at all scales, with few exceptions. While lowering the bit precision generally improves scaling, this trend stops across all models at 3-bit precision, where performance degrades. OPT and Pythia are unstable at 3-bit precision, while GPT-2 and BLOOM remain stable. Plots for all intermediate bit precisions can be found in the Appendix~\ref{appendix:allbits}.}
        \label{fig:scaling_laws}
\end{figure*}

\begin{figure*}[h]
     \centering
         \includegraphics[scale=0.55]{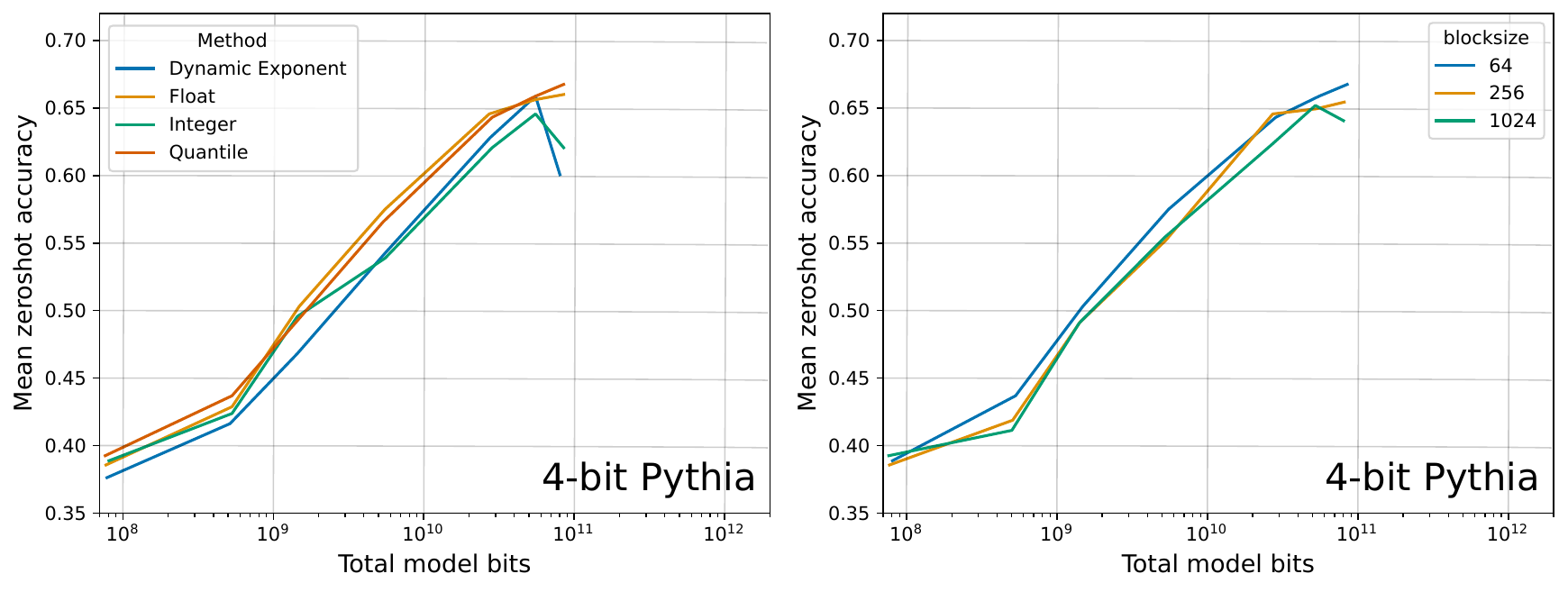}\vspace{-1.0em}
        \caption{Bit-level mean zero-shot scaling laws for 4-bit Pythia models for different data types and block sizes. Block size and data types yield the largest improvements in bit-level scaling. For Pythia, a small block size adds only 0.25 bits per parameter but provides an improvement similar to going from 4-bit to 5-bit precision. In general, across all models, the float and quantile data types yield better scaling than int and dynamic exponent quantization.\vspace{-1em}}
        \label{fig:blocksize}
\end{figure*}

\section{Results \& Analysis}

\subsection{Bit-level Inference Scaling Laws}

The main results are shown in Figure~\ref{fig:scaling_laws}, which depicts the mean zero-shot accuracy over Lambada, PiQA, HellaSwag, and Windogrande, given the total number of bits for OPT, BLOOM, Pythia, and GPT-2 for 3 to 16-bit parameters.

We make the following observations:\vspace{-1em}
\begin{enumerate}
\itemsep=-0em
    \item For a given zero-shot performance, 4-bit precision yields optimal scaling for almost all model families and model scales. The only exception is BLOOM-176B where 3-bit is slightly but not significantly better.
    \item Scaling curves are almost parallel, which indicates that bit-level scaling is mostly independent of scale. An exception to this is 3-bit quantization.
    \item Pythia and OPT are unstable for 3-bit inference where performance is close to random (35\%) for the largest Pythia/OPT models.
\end{enumerate}

\subsection{Improving Scaling Laws}

Given the main scaling results in Figure~\ref{fig:scaling_laws}, an important follow-up question is how we can improve the scaling trends further. 
To this end, we run more than 35,000 zero-shot experiments to vary many of the recently studied advances in quantization precision, including the underlying data types, quantization block size, and outlier-dependent quantization.

These methods usually improve the quantization error at a small cost of additional bits. For example, a block size of 64 with 16-bit quantization constants uses 16 extra bits for every 64 parameters, or 16/64 = 0.25 additional bits per parameter. Outlier-dependent quantization stores the top $p$\% of weight vectors in 16-bit precision and increases the bits per parameter by $p(16-k),$ where $k$ is the precision of the regular weights. For example, for $p=0.02$ and $k=4$, the additional memory footprint is 0.24 bits per parameter.

\textbf{No scaling improvements for 6 to 8-bit precision.} We combine all possible combinations of quantization methods (centering, data types, blocking) with 6 to 8-bit quantization, and we find that none of these methods improve bit-level scaling (see Appendix~\ref{appendix:noscaling}). The main reason for this seems to be that for 6 to 8-bit precision, the model parameters have enough precision to not cause any significant performance degradation compared to 16-bit weights. As such, scaling behavior can only be improved by using less than 6-bit precision rather than enhancing the quantization precision through other means.

\textbf{Small block size improves scaling.} For 3 to 5-bit precision, we do see improvements in scaling by applying quantization methods. Figure~\ref{fig:blocksize} highlights that for 4-bit Pythia models, considerable improvements in bit-level scaling can be achieved by using a small block size. To put this improvement into perspective: Going from a block size of 1024 to 64 adds 0.24 bits per parameter but improves zero-shot accuracy almost as much as going from 4 to 5-bit precision. As such, using a small block size adds a few extra bits compared to improving zero-shot accuracy for 4-bit precision. Besides Pythia, GPT-2 models improve by a large degree. BLOOM, BLOOMZ, and OPT models improve significantly, but less in magnitude compared to Pythia and GPT-2 (see Appendix~\ref{appendix:fourbit-scaling}) -- this relationship likely arises from emergent outlier features. For 5-bit models, the improvement of using small block sizes is minor but still significant. Small block sizes improve 3-bit scaling considerably but still do not make it competitive with 4-bit precision scaling.

\textbf{Data types improve scaling.} From Figure~\ref{fig:blocksize}, we see that data types improve scaling trends for 4-bit Pythia. In particular, the quantile quantization and float data types provide better scaling than integer and dynamic exponent quantization. We generally find that quantile quantization is the best data type across all models, scales, and precisions (see Appendix~\ref{appendix:ppl}). The float data type seems to be superior to Integer quantization with a few exceptions: Integer quantization is better than float quantization for 5-bit -- it appears since the float data type is quite dependent on the balance between exponent and fraction bits, the float data type can be better or worse depending on the particular bit precision. 

 \begin{figure*}[h]
     \centering
         \includegraphics[scale=0.55]{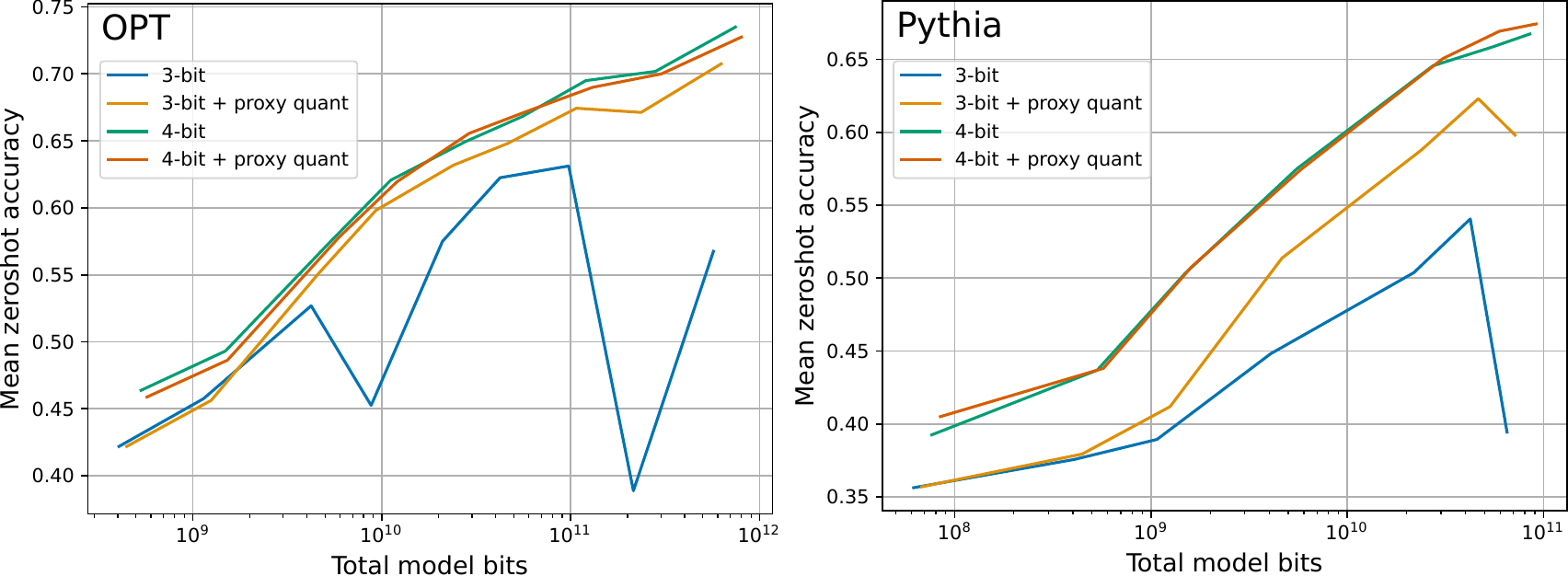}
        \caption{Bit-level scaling laws for outlier dependent quantization for OPT and Pythia. While proxy quantization removes instabilities and improves the 3-bit precision scaling of OPT and Pythia, it still scales worse than 4-bit precision. Proxy quantization is only useful for 3-bit precision weights. Proxy quantization is not useful for models that are relatively stable such as 3-bit BLOOM and GPT-2.}
        \label{fig:outliers}
\end{figure*}

\textbf{Outlier-dependent quantization improves stability, but not scaling.} Finally, we noticed the 3-bit instabilities in OPT and Pythia, and we relate them to emergent outlier features \citep{dettmers2022llm,xiao2022smoothquant}. If we use proxy quantization to quantize the 2\% most significant outlier dimensions to 16-bit instead of 3-bit, we can increase stability and overall scaling for 3-bit Pythia and OPT models. This is shown for OPT in Figure~\ref{fig:outliers}~(left). However, despite this improvement, 4-bit precision still provides better scaling. For 4-bit precision, outlier-dependent quantization has no scaling benefit, as shown in Figure~\ref{fig:outliers}~(right), which means 4-bit precision is optimal despite the considerable improvements for 3-bit precision when using proxy quantization. As such, it appears the outlier features do not require more than 4-bit precision weights to provide optimal bit-level scaling.

\section{Related work}

\paragraph{Large language model quantization.} The most closely related work is on large language model (LLM) quantization for models with more than a billion parameters. Compared to smaller models, LLM quantization poses some unique challenges, such as emergent outliers \citep{dettmers2022llm, zeng2022glm, xiao2022smoothquant} and optimized low-bit inference for LLMs \cite{frantar2022gptq,park2022nuqmm,yao2022zeroquant}. One major defining factor between approaches is zero-shot quantization methods that directly quantize a model without any additional information and one-shot quantization methods that need a mini-batch of data for quantization. While one-shot methods are more accurate, such as GPTQ, which optimizes the rounding during quantization via a mini-batch of data \citep{frantar2022gptq}, they are also more complex and may require hours of optimization before a model can be used. On the other hand, the advantage of zero-shot methods is that they can be used immediately, which makes them easy to use, but zero-shot quantization methods often fail at lower precisions. 

\paragraph{Quantization methods.} Another aspect related to our work are quantization methods which can be grouped into specific categories. For example, there are methods associated with blocking and grouping \cite{park2022nuqmm, wu2020integer, jain2020trained, nagel2019data, krishnamoorthi2018quantizing, rusci2020memory}, centering \citep{krishnamoorthi2018quantizing, jacob2017quantization}, learned data types that are found through clustering \citep{gong2014compressing, han2015deep, choi2016towards, park2017weighted}, or direct codebook optimization \citep{Rastegari2016xnor, hou2016loss, leng2018extremely, zhang2018lq}. While our work studies grouping and blocking, we only study one data type that groups similar weights through their quantiles of the entire input tensor \citep{dettmers2022optimizers}. While we do not study learned data types in depth, we are the first work that shows that these are critical for improving bit-level scaling for LLMs.

\paragraph{Scaling Laws for Inference.} Early work in scaling laws highlighted the importance of studying how variables change with scale since scale is one of the best predictors of model performance \citep{kaplan2020scaling,rosenfeld2019constructive,hestness2017deep}. Particularly, for inference, there has been work that studies scaling trends of zero-shot performance for 4-bit vs. 16-bit models \citep{zeng2022glm}. We study precisions from 3 to 16-bit and disentangle the factors that improve scaling. Work by \citet{pope2022efficiently} looks at scaling inference in a production setting where large batch sizes are common. While they only study quantization rudimentary, they disentangle factors that lead to better model FLOPS utilization (MFU). Since reducing the bit-precision of bits loaded leads to higher MFU, it is similar to our approach to studying bit-level scaling. The main difference is that we vary the bit-width of models and study small batch sizes that are common for consumers and small organizations. 

\section{Recommendations \& Future Work}

We make the following {\bf recommendations}:\vspace{-1em}
\begin{enumerate}
\itemsep=0em
    \item By default, use 4-bit quantization for LLM inference as it offers the total model bits and zero-shot accuracy trade-offs.
    \item Use a block size of 128 or lower to stabilize 4-bit quantization and improve zero-shot performance.
    \item Use a floating point or quantile quantization data type. In some cases, integer data types might be preferable to improve inference latency depending on the implementation and hardware support.
\end{enumerate}

A case where a higher than 4-bit precision is desirable is when one works with a GPU with enough memory to hold a higher bit precision but not a larger model. For example, a 48 GB GPU has enough memory to use a 66B model in 5-bit precision but cannot fit a 175B model in 4-bit. Therefore, if maximal zero-shot accuracy is desired, 5-bit precision and a 66B model is preferable for this scenario.

\begin{table}[]
\centering\vspace{-0em}
\caption{WikiText-2 perplexity for 2-bit GPTQ and 3-bit Float with blocking. We can see that GPTQ is superior to 3-bit Float if blocking is used. As such, methods like GPTQ are a promising way to improve low-bit scaling. However, GPTQ requires blocking to provide good scaling (see Figure~\ref{fig:gptq}).}
\label{tbl:gptq}
\begin{tabular}{ccc}\toprule
\multicolumn{1}{l}{} & \multicolumn{2}{c}{WikiText-2 Perplexity} \\\cmidrule{2-3}
\multicolumn{1}{l}{Blocksize} & \multicolumn{1}{l}{2-bit GPTQ} & \multicolumn{1}{l}{3-bit Float} \\\midrule
1024 & {\bf 11.84} & 13.26 \\
256 & {\bf 10.00} & 10.38 \\
64 & {\bf 9.18} & 9.99 \\\bottomrule\vspace{-1em}
\end{tabular}
\end{table}

\begin{figure}[]
     \centering
         \includegraphics[scale=0.55]{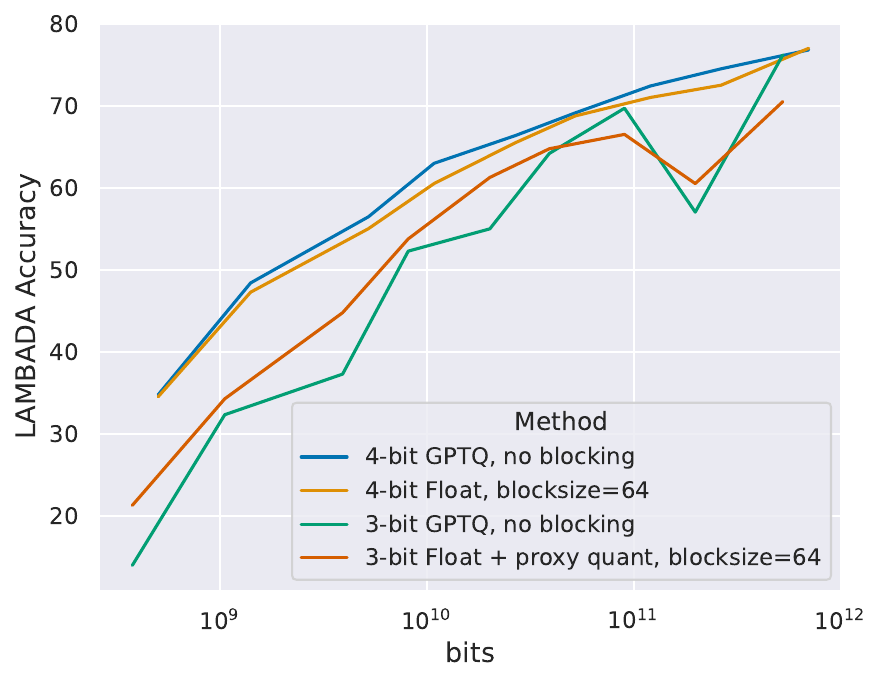}\vspace{-1.5em}
        \caption{Bit-level scaling laws for LAMBADA zero-shot accuracy. We can see that GPTQ without blocking scales poorly at 3-bit. While one-shot quantization methods like GPTQ are superior to zero-shot methods like proxy quantization, other methods like blocking are required to make them bit-level efficient (see Table~\ref{tbl:gptq}).\vspace{-1em}}
        \label{fig:gptq}
\end{figure}

{\bf Promising directions for future work}. Our results highlight that 4-bit precision is currently bit-by-bit the most efficient precision, but we also show that 3-bit scaling can be significantly improved. As such, one major promising research direction is to focus on low-bit precisions below 4-bit and improve their scaling trends. While our methods are zero-shot quantization methods, meaning they do not use any input data or optimization procedure to optimize the quantization, it has been shown that one-shot quantization methods, like GPTQ, are more effective at low-bit precisions \citep{frantar2022gptq}. Table~\ref{tbl:gptq} shows that 2-bit GPTQ with blocking yields better performance than zero-shot 3-bit Float. This highlights that one-shot methods are very promising for low-bit precisions. 

On the other hand, Figure~\ref{fig:gptq} also shows that 3-bit GPTQ without blocking scales worse compared to 3-bit Float with a blocksize of 64 and 4-bit GPTQ without blocking yields similar scaling compared to 4-bit Float with a blocksize of 64. From these results, it appears that the insights gained from our zero-shot scaling laws translate to one-shot quantization methods. This shows that zero-shot quantization research is well suited for disentangling and understanding individual factors in quantization scaling, while one-shot methods maximize performance. In contrast, current one-shot methods are expensive to study. For example, repeating our scaling experiments for GPTQ only for the OPT-175B and BLOOM-176B models would consume an estimated 5,120 GPU days of compute. Therefore, the combination of zero-shot quantization scaling insights and one-shot quantization methods may yield the best future methods.

One challenge unaddressed in our work is that the data type should also be able to be used efficiently on a hardware level. For example, while quantile quantization is the data type that shows the best scaling trends, it requires a small lookup table that is difficult to implement in a highly parallel setup where parallel memory access often leads to poor performance due to the serialization of threads. This problem can be solved by designing new data types that are both bit-level scaling efficient and hardware efficient. Another perspective to solve this problem could be hardware design: Can we design hardware to use data types such as quantile quantization efficiently?

Both block-size and outlier-dependent quantization improve the quantization precision of outliers. While outlier-dependent quantization does not offer improvements in scaling, it is reasonable that there are unknown quantization methods that help with outliers and improve scaling trends simultaneously. As such, another primary focus of future research should center around preserving outlier information while minimizing the use of additional bits.

\section{Discussion \& Limitations}

While we ran more than 35,000 experiments, a main limitation is that we did not consider certain classes of quantization methods. For example, there are quantization methods where a data type is optimized with additional input data \citep{Rastegari2016xnor,frantar2022gptq} or from the weights of the model alone \citep{gong2014compressing}. Optimization from the weights alone is similar to quantile quantization which was the most effective data type in our study. As such, this hints that such quantization methods could improve scaling for inference and present a missed opportunity to be included in our study. However, our study is an essential step towards recognizing the importance of these methods for optimizing the model-bits-accuracy trade-off -- a perspective that did not exist before.

Another limitation is the lack of optimized GPU implementations. It is unclear if other data types that rely on lookup tables can achieve significant speedups. However, efficient implementations for our Int/Float data types would be possible, and our results for other data types are still useful for the development of future data types, which yield strong scaling and efficient implementations.

While we only study the latency-optimal perspective indirectly through studying the model-bits-accuracy trade-off, a practical limitation of the latency-optimal perspective is that low-bit models with 16-bit inputs might be less latency efficient if such a model is deployed to be used by many users \citep{pope2022efficiently}. Given a busy deployed API with thousands of requests per second, the large mini-batches would no longer fit into the cache. This means bit-level scaling laws would be increasingly unrelated to inference latency in this case. For such high throughput systems, scaling laws that model both low-bit weights {\it and} low-bit inputs are required to study optimal inference latency scaling. In short, our scaling laws are only valid for cases where the mini-batch does not fit into the L1 cache of the device, and beyond this, a new set of scaling laws is required.

A final limitation is that loading the weight matrix is only one part of inference latency which needs to be optimized to achieve fast inference. For example, without optimizations for the attention operations, multi-head attention can be a large chunk of the inference latency footprint \citep{jaszczur2021sparse,pope2022efficiently}. However, the overall memory footprint of the model is still reduced, making large language models more easily usable when GPU memory is limited. 

\section{Conclusion}

Here we presented a large-scale study of 35,000 zero-shot experiments on a wide variety of LLMs and parameter scales to analyze the scaling behavior and trade-offs between the number of parameters, quantization bit precision and zero-shot accuracy during inference. We find that 4-bit quantization is almost universally optimal to reduce the model bits and maximize zero-shot accuracy. We study the improvement of bit-level scaling behaviors and find that data types and block size are the most critical measures to improve bit-level scaling. Our analysis paves the way for the systematic study of inference scaling trends for LLMs.



\newpage

\bibliography{example_paper}
\bibliographystyle{icml2022}

\newpage
\appendix
\onecolumn

\section{Data type details}
\label{appendix:datatypes}

These sections provide the full details of all our data types that we used in our k-bit quantization experiments. First, we provide a general view of quantization that unifies data types under a common formalism so that data types can be more easily compared. We discuss the limitations of this view in our discussion section. 

\paragraph{Quantization as a mapping from integers to values.} While there are many ways to define quantization, we provide a general definition that unifies data types. We define quantization as a mapping from $k$-bit integers $I$ to floating point values $F$ in the range $[-1, 1]$. This definition has the advantage that a data type is fully specified by the set of floating point values $F$ alone and the number of bits $k$ in the set $I$.

More formally, we can describe $k$-bit quantization as a mapping from the set of $k$-bit integers $I$ to the set $F$, that is, ${\mathbf{Q}^{\text{map}}: I \mapsto F = [0, 2^k-1] \mapsto F}$. For example, the IEEE 32-bit floating point data type maps the indices $0...2^{32}-1$ to the set $S$ with domain \mbox{[-3.4e38,~+3.4e38]}. Furthermore, to be able to compare data types more easily, we normalize the domain of the real values in set $F$ to the range $[-1, 1]$ by dividing by the absolute maximum value of the set of possible values $F$. We use the following notation: ${\mathbf{Q}_k^{\text{map}}(i) = q_i}$, for example ${\mathbf{Q}_{8}^{\text{uint}}(83) = 83/255 = 0.3255}$ for an 8-bit unsigned integer data type. With this notation, the mapping index $i$ represents the quantized value to be stored.

\textbf{To quantize an arbitrary input}, we normalize the input into the range $[-1, 1]$\footnote{This range is used for storage and not computation.} and then do a binary search in set $F$ to find the closest real value to the input and its associated mapping index $i$. Once the closest value is found, we store the quantized value as the mapped index. Formally, this can be described as:

\begin{equation}
    \mathbf{T}_i^{Q_k^{\text{map}}} = \argmin\limits_{j=0}^{n=2^k} |\mathbf{{Q}}_k^{\text{map}}(j) - {\mathbf{T}_i}|,
\end{equation}

where $\mathbf{T}$ is the normalized input tensor. 

\textbf{To dequantize} the tensor $\mathbf{T}_i^{Q_k^{\text{map}}}$ back to a real-valued tensor $\mathbf{T}^F$, we perform a lookup:

\begin{equation}
\mathbf{T}^F_i = {\mathbf{Q}^{\text{map}}}(\mathbf{T}^Q_i)\cdot c,
\end{equation}

where $c$ is the normalization constant that normalized $\mathbf{T}$ into the range $[-1, 1]$.

\textbf{To do computation}, we dequantize the weight in the cache and perform a 16-bit floating point multiplication with the 16-bit input.

With this general definition, we can now define the following data types by defining their set of quantization values $F$, also called a codebook.

\paragraph{Integer data types.} Integer quantization, also known as linear or uniform quantization, maps the integers to itself with an offset of 128 for a signed integer ${\mathbf{Q}^{\text{int}}: I \mapsto F = [0, 2^k-1] \mapsto [-(2^{k-1}-1), 2^{k-1}]}$. In practice, we truncate the set $F$ to have an equal number of positive and negative values around zero. So, for example, for an Int8 data type, we have the values $[-127/c, 127/c]$ where $c=127$ is the absolute maximum normalization constant. 

\paragraph{Floating Point data types.} Floating point data types are represented by a combination of exponent bits E (with base 2) and mantissa bits M (fraction). Since we use 3-8 bit precision in our work, we take the FP8 data type as a reference \citep{micikevicius2022fp8}. The only difference is that we do not allocate a value for NaN values. As such, the floating point data type we use is defined by the following equations:

\begin{equation}
   \begin{array}{c}
   \text{Subnormal numbers, if the exponent is zero:} \\
    (-1)^{\text{signbit}} \times 2^{-\text{bias}} \times \sum\limits_{i=1}^{\text{M}} b[\text{M}-i]\cdot2^{-i}  \\
   \text{Normalized numbers, if the exponent is non-zero:} \\
   (-1)^{\text{signbit}} \times 2^{-({\text{E}-1-\text{bias}})} \times (1 + \sum\limits_{i=1}^{\text{M}} b[\text{M}-i]\cdot2^{-i}) 
   \end{array},
\end{equation}

where bias=$2^{E-1}+1$, signbit=$\{0, 1\}$, and $b[i]$ represents the binary mantissa bit value at index $i$ in the bit-mask. To create the set $F$, we can iterate through all possible combinations of exponent, mantissa, and sign bits and apply the equations above. We evaluate different combinations of exponent and mantissa bits and found that a 2 to 3-bit exponent performs best in terms of zero-shot accuracy. We used a 3-bit exponent for 4 to 8-bit precision quantization, a 2-bit exponent for 3-bit quantization. As such our 4-bit and 5-bit Float results are slightly sub-optimal. For more on how different exponent bit combinations relate to performance see Appendix~\ref{appendix:float}.

\paragraph{Dynamic Exponent.} Dynamic exponent data types \citep{dettmers20168bit,dettmers2022optimizers} use one bit for the sign, and the number of following zero bits represents the exponent with base 10. The first bit, which is one, serves as an indicator bit that separates the exponent and the unsigned linear fraction. The remaining bits represent an unsigned linear quantization. Alternatively, we can construct the set of floating point values $F$ for the dynamic exponent data type by bisecting the interval $[0.1, 0.9]$ into $n$ equal intervals where $n$ is the number of fractional bits. We also define $00000000_2 = 0_{10}$ -- which means we add the value of zero to $F$.

\begin{figure}
  \centering
  \includegraphics[scale=0.25]{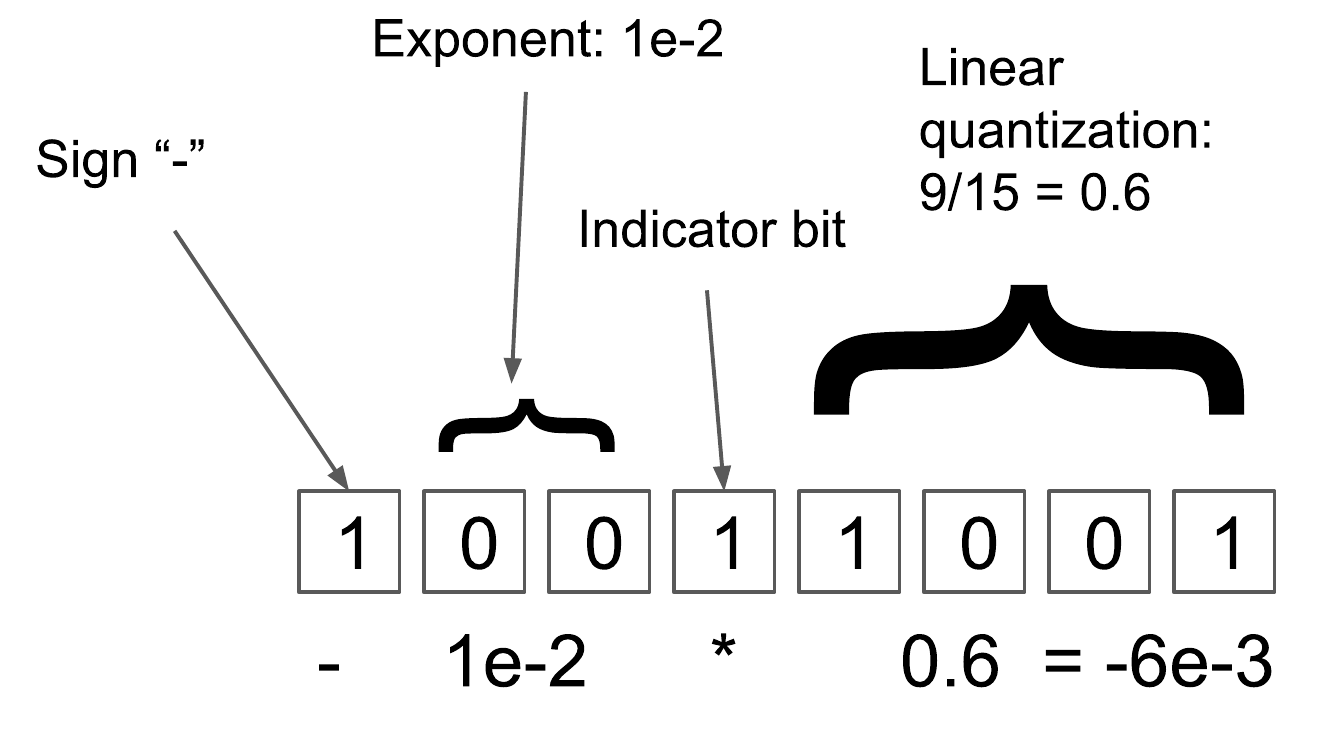}
  \caption{Schematic of dynamic exponent quantization.}
\label{fig:dynamic_exponent}
\end{figure}

\paragraph{Quantile Quantization.} The information-theoretically optimal data type will allocate an equal number of values to each quantization bin. For example, for a 1-bit quantization, this would be a quantization map with values $F$ containing the lower and upper quartile so that 50\% of values are allocated to each of the two bins. Quantile quantization \cite{dettmers20168bit} describes this information-theoretically optimal data type for a k-bit quantization. It can be defined as such:

\begin{equation}
q_i  = \dfrac{Q_X\left(\frac{i}{2^k+1}\right) + Q_X\left(\frac{i+1}{2^k+1}\right)}{2},
\end{equation}

where $Q_X$ is the quantile function which is the inverse of the cumulative distribution function $Q_X = F_X^{-1}$. We estimate the values of the quantile function by using the SRAM Quantiles algorithm \citep{dettmers2022optimizers}, which approximates the quantiles of an input tensor through the empirical cumulative distribution function of the tensor. The set $F$ is defined by all $q_i$ from 0 \dots $2^k-1$, and an additional 0 is added to the set $F$.

\section{Further negative results: distribution centering}

Here we use the notation introduced in Appendix~\ref{appendix:datatypes} to define distribution centering and present negative results.

In \textbf{distribution centering}, we subtract the mean from the input tensor before quantization so that asymmetric distributions are approximately centered around zero. 

\begin{equation}
\begin{array}{c}
    m = \sum\limits_{i=0}^n \mathbf{T}_i/n \\
    \mathbf{T}_i^{Q_k^{\text{map}}} = \argmin\limits_{j=0}^{n=2^k} |\mathbf{{Q}}_k^{\text{map}}(j) - ({\mathbf{T}_i}-m)|
    \end{array}
\end{equation}

We add the mean as the final operation to dequantize a distribution-centered value.

\begin{equation}
\mathbf{T}^F_i = {\mathbf{Q}^{\text{map}}}(\mathbf{T}^Q_i)\cdot c + m,
\end{equation}

\textbf{Scaling results: Distribution centering is ineffective.} While it has been shown that for activations that centering the distributions around zero can improve quantization errors due to asymmetric distributions (such as ReLU outputs), we find that distribution centering does not improve scaling for weight quantization in any scenario.

\section{Detailed Scaling Results}

\subsection{Full Scaling Laws for 3-bit to 16-bit}
\label{appendix:allbits}

\begin{figure*}[t]
     \centering
         \includegraphics[scale=0.55]{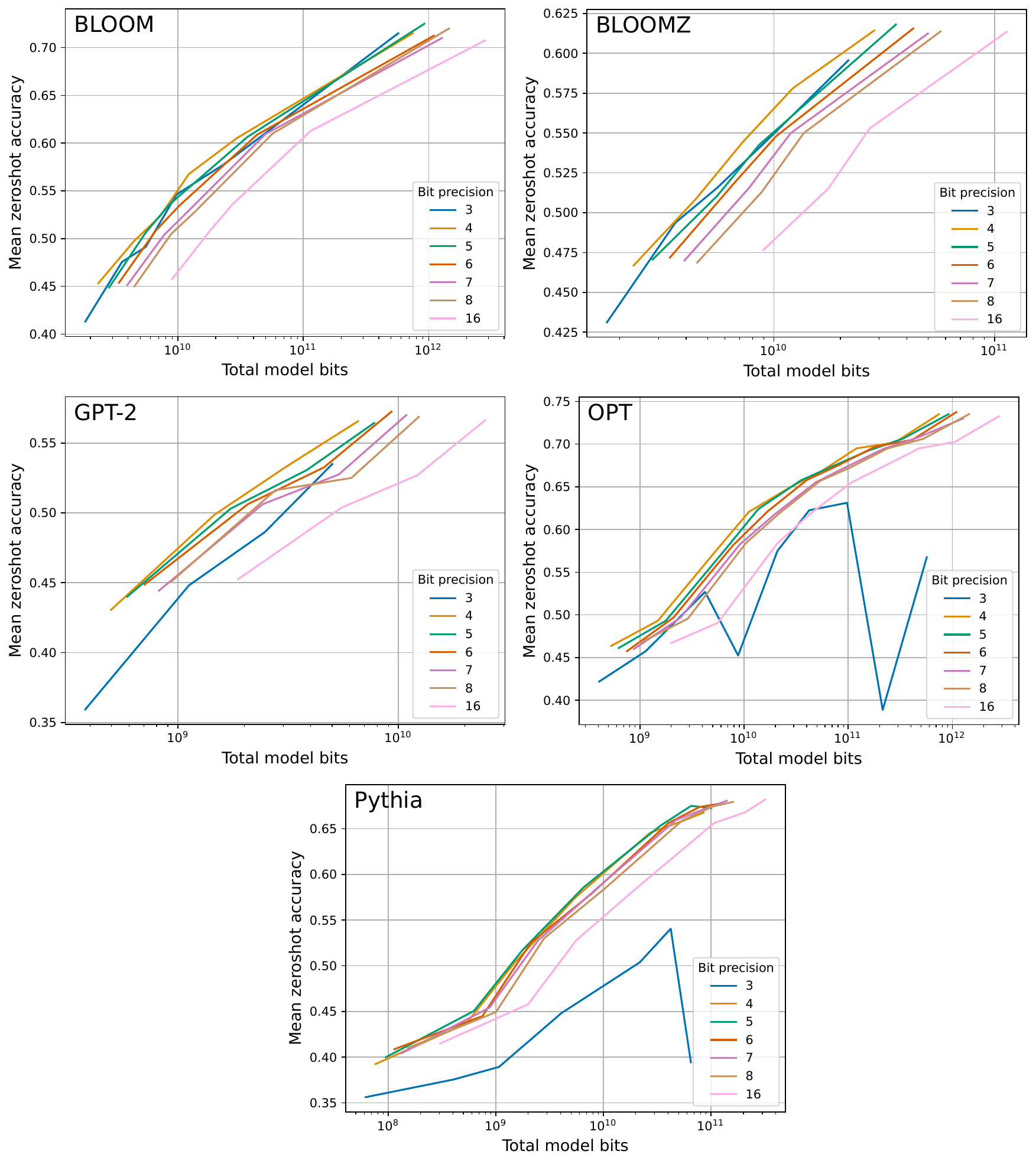}
        \caption{Bit-level scaling laws for mean zero-shot performance across Lambada, PiQA, Winogrande, and Hellaswag. 4-bit precision is optimal for all models at all scales with few exceptions. While lowering the precision generally improves scaling, this trend stops across all models at 3-bit precision where performance degrades. OPT and Pythia are unstable at 3-bit precision while GPT-2 and BLOOM remain stable. Stability is related to outlier features as explored in our analysis section.}
        \label{fig:scaling_laws_full}
\end{figure*}

Figure~\ref{fig:scaling_laws_full} shows bit-level scaling from 3 to 16-bit precision. Notable exceptions not found in the main paper scaling trends in these plots are as follow:
\begin{enumerate}
    \item Pythia 5-bit as good as Pythoa 4-bit.
    \item BLOOM and BLOOMZ show almost the same quantization behavior, indicating that fine-tuning an existing model does not change its quantization properties.
\end{enumerate}

\subsection{Details for scaling improvements at 4-bit precision}
\label{appendix:fourbit-scaling}

\begin{figure*}[t]
     \centering
         \includegraphics[scale=0.55]{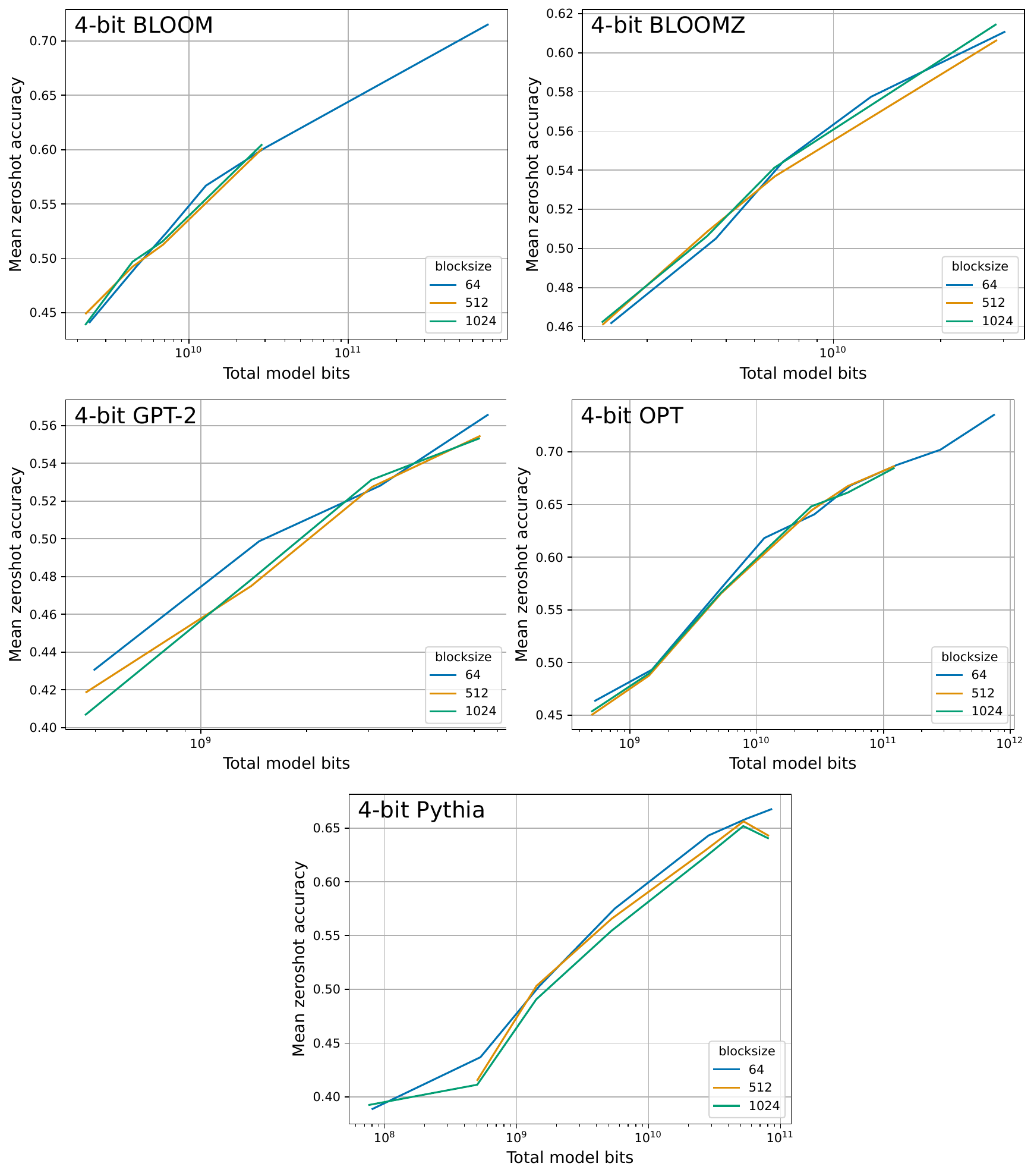}
        \caption{4-bit scaling laws for mean zero-shot performance across Lambada, PiQA, Winogrande, and Hellaswag for different quantization data types. We see that the choice of block size improves bit-level scaling for most models at most scales. But this relationship is not always straightforward as in the case of OPT. We see more clear trends for 3-bit precision or when comparing methods using perplexity (not shown). }
        \label{fig:fourbit-blocksize}
\end{figure*}

\begin{figure*}[t]
     \centering
         \includegraphics[scale=0.55]{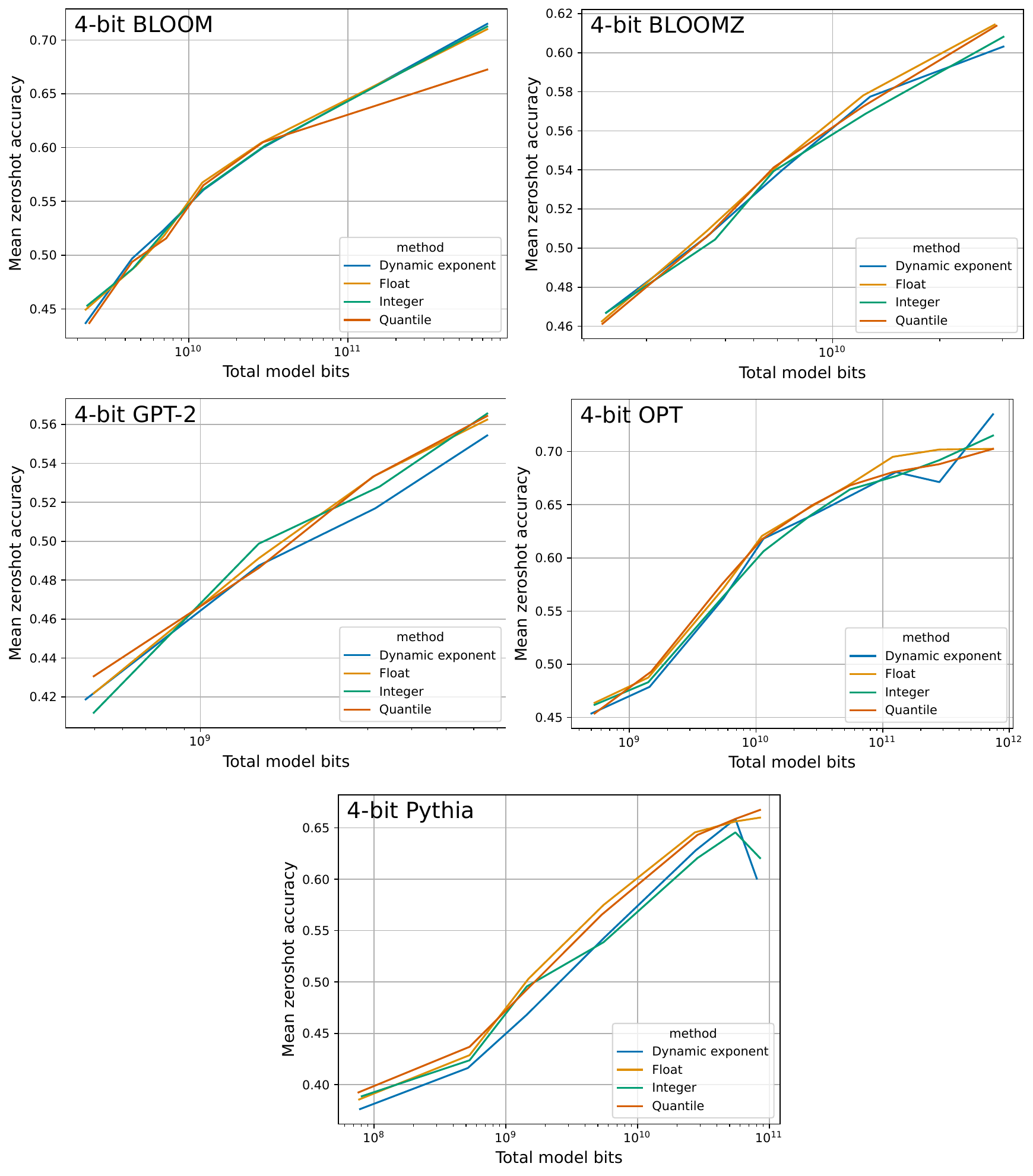}
        \caption{4-bit scaling laws for mean zero-shot performance across Lambada, PiQA, Winogrande, and Hellaswag for different quantization data types. We see that the choice of data types improves bit-level scaling for most models at most scales with quantile quantization provided the best scaling on average. }
        \label{fig:fourbit-method}
\end{figure*}

The main improvements that we see through quantization methods occur at 3-bit and 4-bit precision. Here we give full details for the 4-bit scenario. Figure~\ref{fig:fourbit-blocksize} shows how bit-level scaling can be improved by using a smaller blocksize. Figure\ref{fig:fourbit-method} shows bit-level scaling for different data types. We did not perform a grid search on blocksize for 4-bits for BLOOM-176B and OPT-175B since we lacked the compute to complete these experiments.

\subsection{No scaling improvements for 6 to 8-bit models through quantization methods}
\label{appendix:noscaling}

In this section we present data that shows that we cannot improve the bit-level scaling if we add quantization techniques that improve quantization precision if we use 6 to 8 bits per parameter. We hypothesize that this is because 6 to 8 bits per parameter is sufficient to model the weights with enough precision to not cause any major quantization precision problems. For example, any outliers in the weights might be sufficiently modelled by 6 to 8 bits and do not require additional techniques to prevent major quantization errors. 

\begin{figure*}[t]
     \centering
         \includegraphics[scale=0.55]{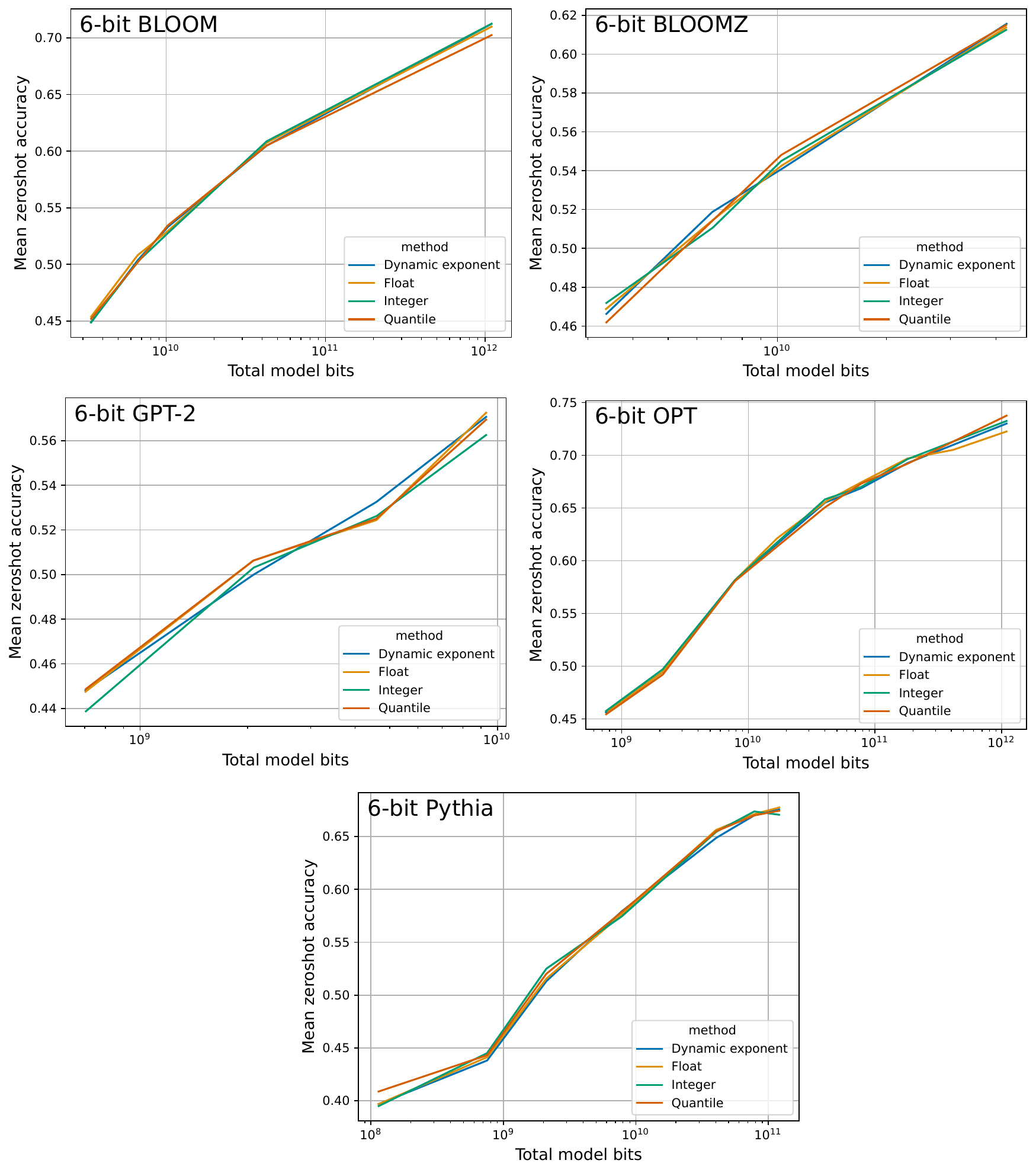}
        \caption{6-bit scaling laws for mean zero-shot performance across Lambada, PiQA, Winogrande, and Hellaswag for different quantization data types. We see that the choice of data types does not affect scaling behavior at 6-bit precision. We results are similar tor 7 and 8-bit precision. }
        \label{fig:sixbit-method}
\end{figure*}

\begin{figure*}[t]
     \centering
         \includegraphics[scale=0.55]{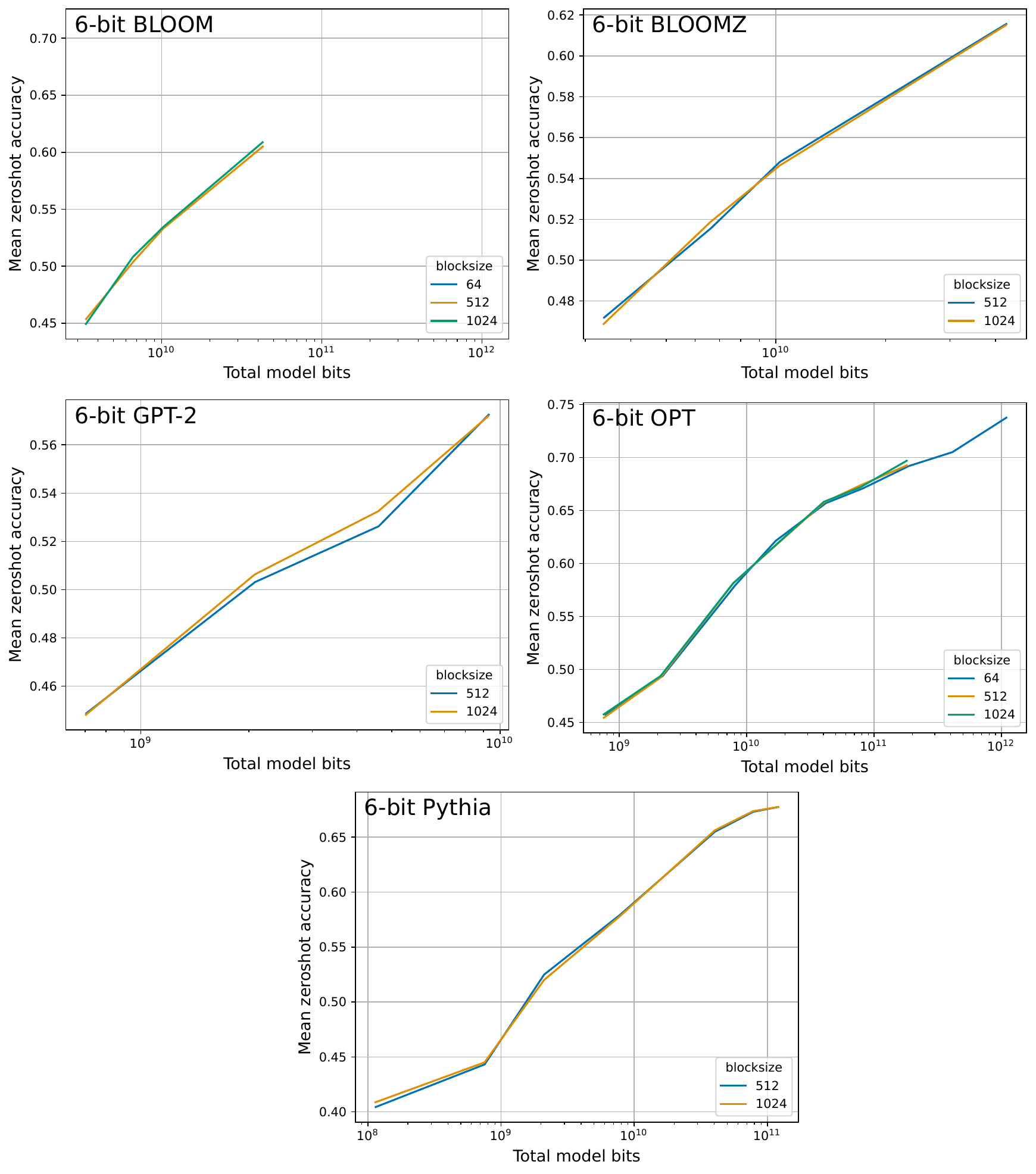}
        \caption{6-bit scaling laws for mean zero-shot performance across Lambada, PiQA, Winogrande, and Hellaswag for different quantization data types. We see that the choice of the blocksize does not affect scaling behavior at 6-bit precision. We results are similar tor 7 and 8-bit precision. }
        \label{fig:sixbit-blocksize}
\end{figure*}

Figure~\ref{fig:sixbit-method} shows that we cannot improve bit-level scaling through data types when using 6-bit per paramters. We find similar results for 7 and 8-bit precision. Figure~\ref{fig:sixbit-blocksize} shows a similar relationship for the block size variable.

\subsection{Scaling of floating point exponent bits}
\label{appendix:float}

It has been studied before how the standard deviation of the input relates to the optimal exponent bit configuration for the FP8 data type \citep{jin2022f8net}. However, this format assumes row-wise quantization without any blocking of the weight. Its also not studied how the overall performance is affected as we scale.

\begin{figure*}[t]
     \centering
         \includegraphics[scale=0.55]{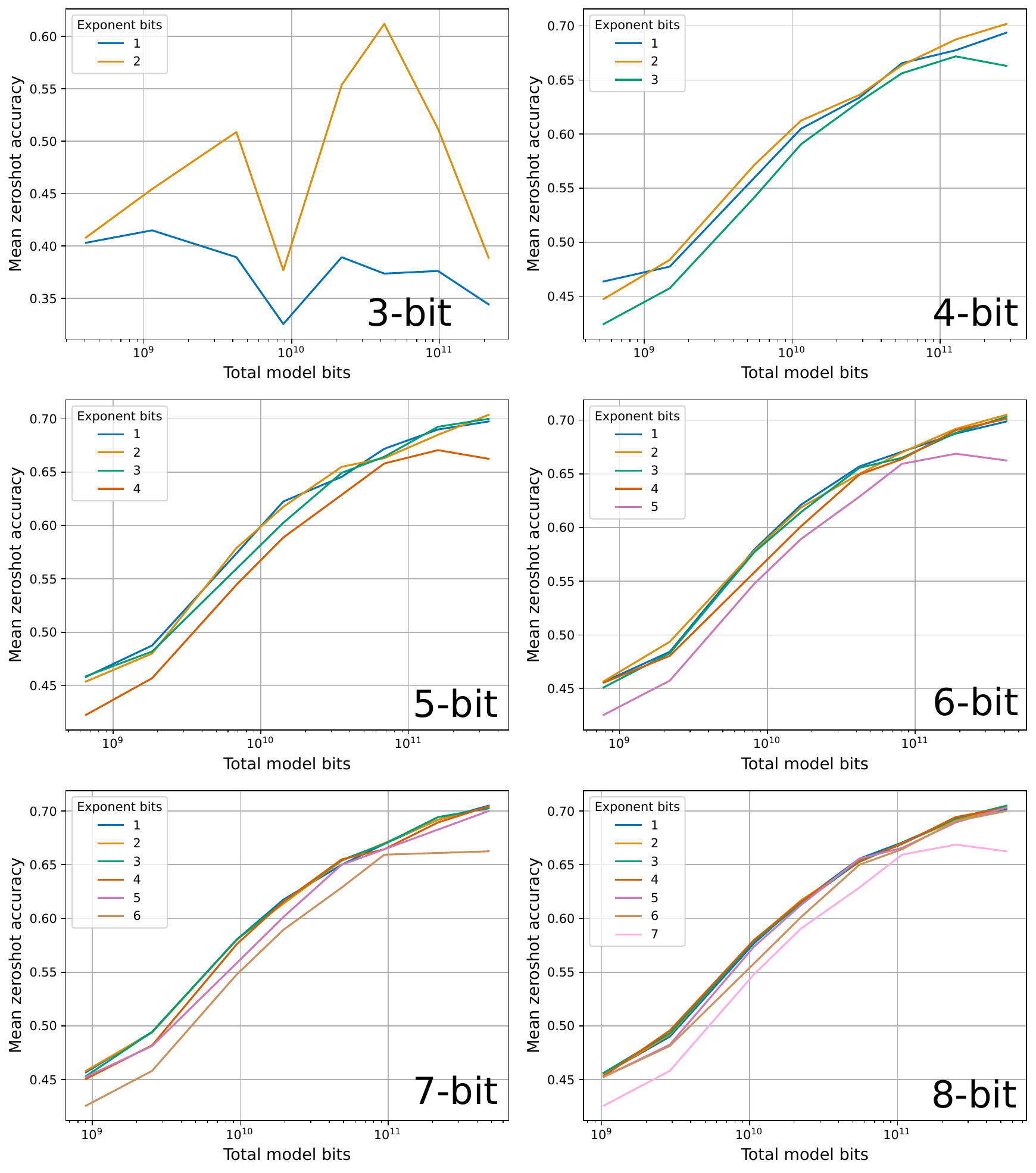}
        \caption{Inference scaling laws for mean zero-shot performance across Lambada, PiQA, Winogrande, and Hellaswag for different exponent bits for different bit precisions for the float data type and block-wise quantized weight. We see that a 2-bit exponent is the only one that performs well across all precisions.  }
        \label{fig:ebits}
\end{figure*}

Here we present data on the scaling behavior of 3 to 8-bit precision quantization on transformer performance as we scale OPT from 125M to 176B. We use block-wise quantization with block wise 64 for the weight. Figure~\ref{fig:ebits} shows that float data types with relatively many exponent bits do well if we have row-wise quantized inputs and block-wise quantized weights. A good heuristic is that for any bit-precision, the exponent bits should make up at least half the bits rounded up. This means, for 3, 4, 5, 6, 7, 8 bits we should use 2, 2, 3, 3, 4, 4 exponent bits. The only precision where this heuristic is not optimal is for 5-bit precision.

\subsection{Perplexity-based inference scaling laws}
\label{appendix:ppl}

\begin{figure*}[t]
     \centering
         \includegraphics[scale=0.55]{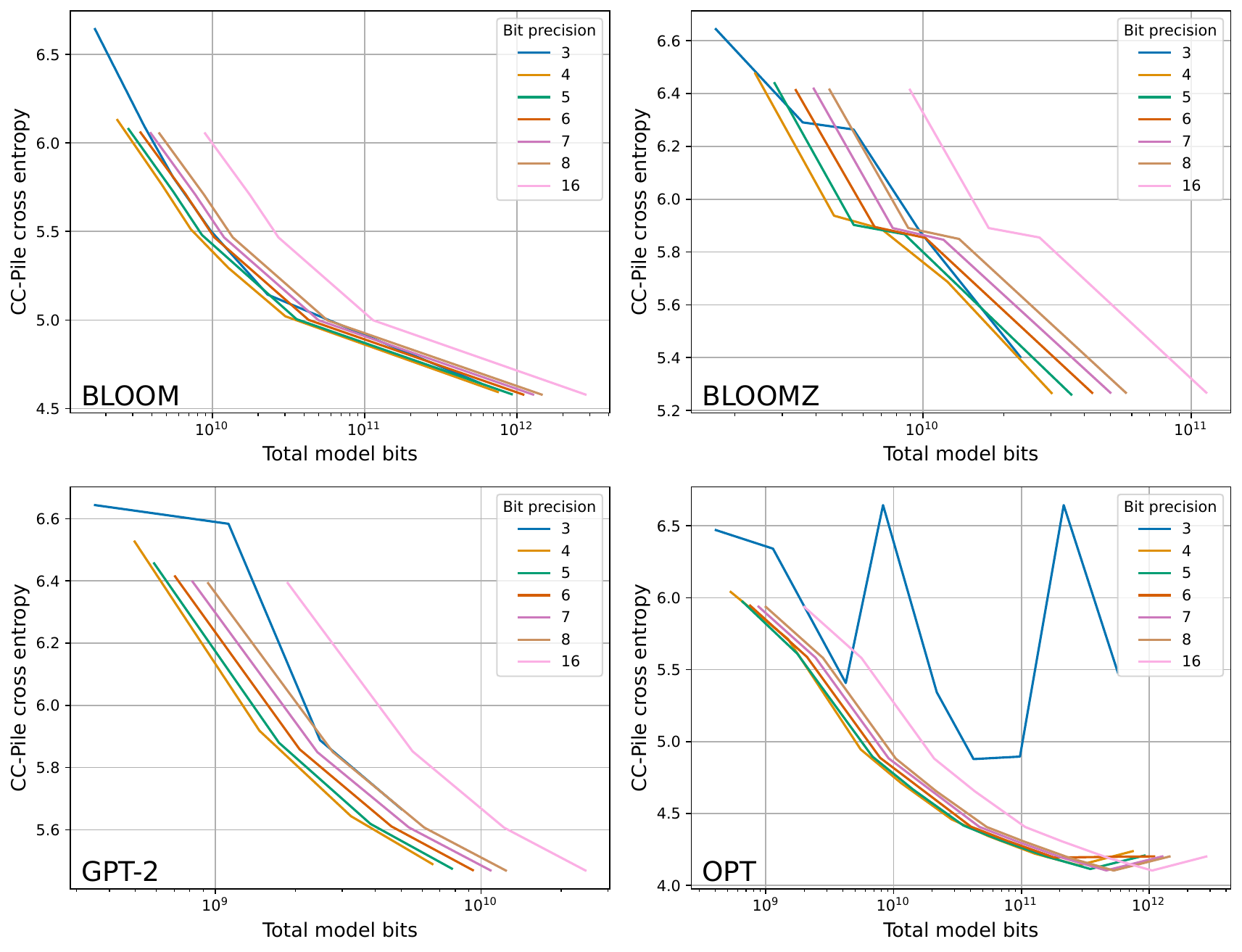}
        \caption{Inference scaling laws for cross entropy loss on CC-Pile for different bit precisions. We see that 4-bit quantization is bit-level optimal across all models.  }
        \label{fig:ppl-bits}
\end{figure*}

\begin{figure*}[t]
     \centering
         \includegraphics[scale=0.55]{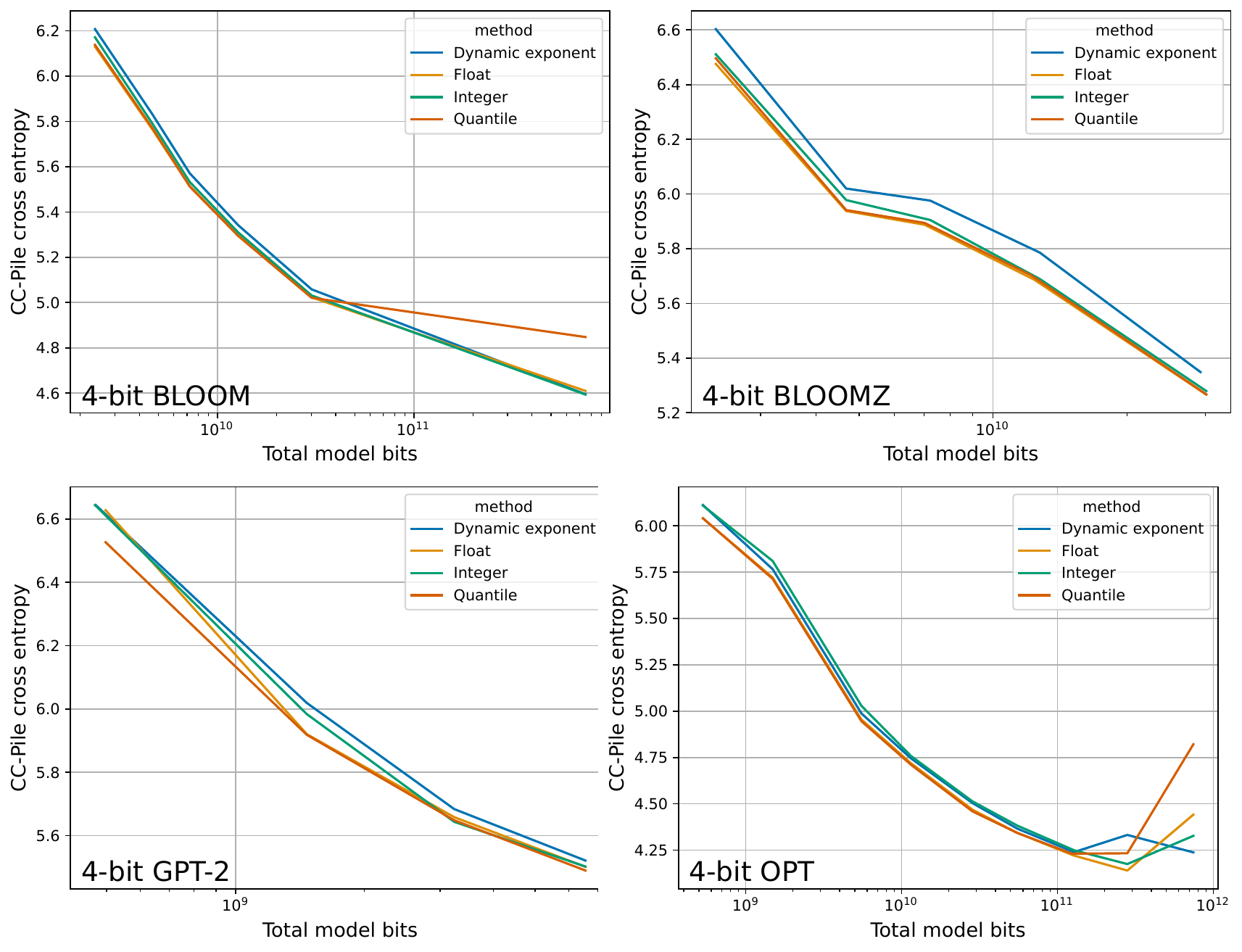}
        \caption{Scaling laws for cross entropy loss on CC-Pile for different quantization data types. We see that quantile quantization is the best data type on average. }
        \label{fig:ppl-method}
\end{figure*}

\begin{figure*}[t]
     \centering
         \includegraphics[scale=0.55]{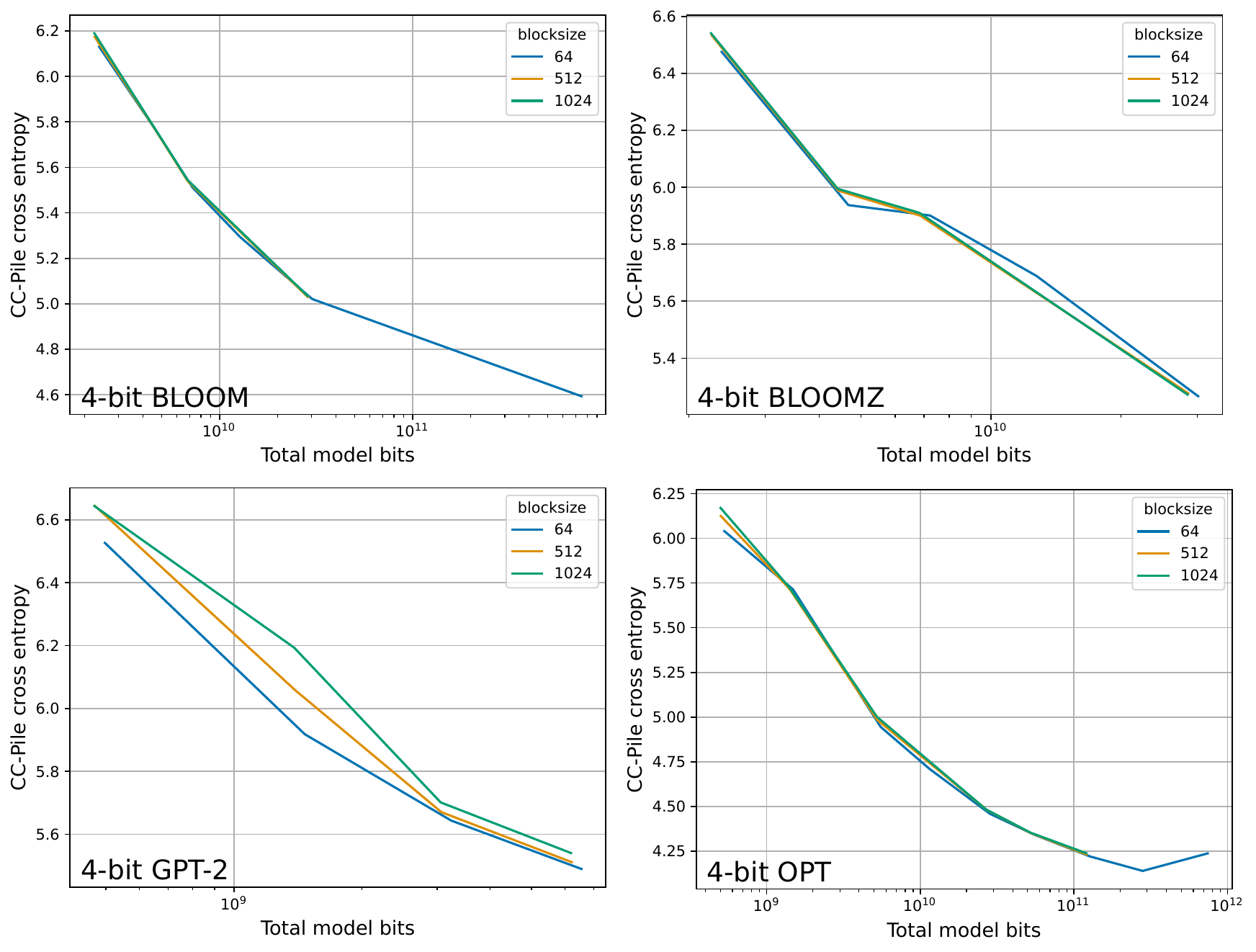}
        \caption{Inference scaling laws for cross entropy loss on CC-Pile for different block sizes. We see that a smaller block size is better than a larger one. While the difference is small across many models, small block sizes are consistently better, except for BLOOMZ.}
        \label{fig:ppl-blocksize}
\end{figure*}

Since perplexity evaluation provides a continuous value with each token and zero-shot accuracy only a binary value, perplexity is the more reliable value to compare methods. In this section, we present data for evaluation on The Pile Common Crawl \citep{gao2020pile}. We did not evaluate Pythia models since our codebase had a bug where evaluation crashed and only provide evaluation for BLOOM, BLOOM, GPT-2, and OPT. For OPT-175B we used less samples to evaluate since we did not have enough compute to produce these metrics before the conference deadline and this introduced noise which makes perplexity for OPT-175B worse than OPT-66B. In our graphs we limit the perplexity to 100 which indicates that the quantization was unstable and performed at random performance (perplexity $>$ 100000). We also transform the perplexity to the cross entropy loss value to make curves more easily visibile. Figure~\ref{fig:ppl-bits} shows inference scaling laws for total bits, Figure~\ref{fig:ppl-method} for data types, and Figure~\ref{fig:ppl-blocksize} for block size.

\end{document}